\pdfoutput=1
\documentclass[11pt]{article}

\usepackage{acl}

\usepackage{times}
\usepackage{latexsym}
\usepackage[autostyle]{csquotes}
\usepackage{float}

\usepackage[T1]{fontenc}
\usepackage[utf8]{inputenc}
\usepackage{graphicx} 
\usepackage{tabularx,ragged2e}
\usepackage{array, makecell}
\newcolumntype{C}{>{\Centering\arraybackslash}X} % centered "X" column

\usepackage{microtype}

\usepackage{inconsolata}

\title{    Diving Deep into the Motion Representation of Video-Text Models}
 \author{ Chinmaya Devaraj \quad   Cornelia Ferm\"uller \quad   Yiannis Aloimonos \\
         University of Maryland, College Park  \\ (chinmayd,fermulcm,jyaloimo)@umd.edu }
\begin{document}
\maketitle

\begin{figure*}[t]

   \includegraphics[width=0.9\linewidth]{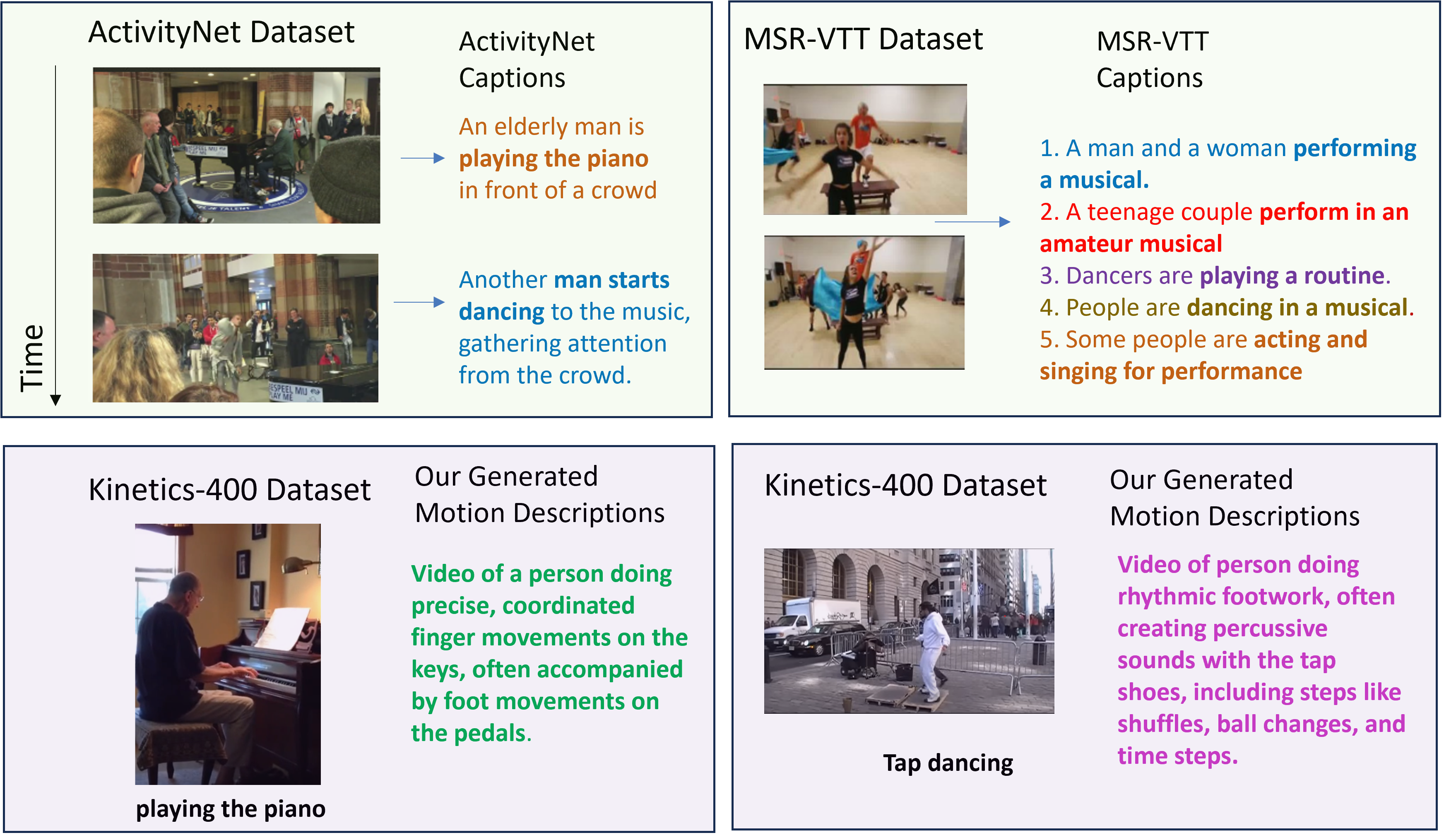}
  \centering
\caption{ Example captions from ActivityNet, MSR-VTT, and our own GPT-4 generated fine-grained motion description for Kinetics-400 classes. Our generated motion descriptions solely describe the motion of the action, whereas other datasets typically use verbs to describe the scene. }
\label{fig:datasets}
\end{figure*}

\begin{abstract}
Videos are more informative than images because they capture the dynamics of the scene. By representing motion in videos, we can capture dynamic activities. In this work, we introduce GPT-4 generated motion descriptions that capture fine-grained motion descriptions of activities and apply them to three action datasets. We evaluated several video-text models on the task of retrieval of motion descriptions. We found that they fall far behind human expert performance on two action datasets, raising the question of whether video-text models understand motion in videos. To address it, we introduce a method of improving motion understanding in video-text models by utilizing motion descriptions.  This method proves to be effective on two action datasets for the motion description retrieval task. The results draw attention to the need for quality captions involving fine-grained motion information in existing datasets and demonstrate the effectiveness of the proposed pipeline in understanding fine-grained motion during video-text retrieval. 

\end{abstract}

\section{Introduction}
Since the introduction of large-scale use of contrastive learning for image and text representation~\citep{radford2021learning}, various efforts have been made to build video-text models \citep{ni2022expanding, luo2022clip4clip,fang2021clip2video, wang2021actionclip} to relate video to text. Videos provide a way to access the dynamics or motion in the scene that a single image cannot capture~\citep{fermuller2018prediction, fermuller2022learning, dessalene2023therbligs}. Motion in videos could be due to the action depicted, the effect of camera movement (for example, in egocentric action videos), or a combination of camera motion and action~\citep{ogale2005motion}. 

%\begin{figure}[t!]

%   \includegraphics[width=0.9\linewidth]{latex/teaser_fig1.png}
%  \centering
%\caption{\textbf{Qualitative examples of the additional descriptions of verbs :}  The left side shows the video clip and original caption. The right side box shows the video's motion description generated by GPT-4.}
%\label{fig:dataset}
%\end{figure}

We investigate how existing video-text models perceive motion due to the action. For this work, we define motion in action videos as the movement of actors or the movement of actors and objects. The challenge is the lack of datasets that explicitly describe video motion. Figure \ref{fig:datasets} shows some examples of captions from ActivityNet \citep{krishna2017dense} and the MSR-VTT \citep{xu2016msr} dataset.  Although verbs are included in captions of multimodal datasets like ActivityNet, MSR-VTT, Howto100M \citep{miech2019howto100m}, Spoken Moments in Time \citep{monfort2021spoken}, a detailed description of motion is not available. This calls for the need to have an exclusive benchmark to evaluate how video-text models interpret and respond to motion descriptions.

%Since motion is very much interlinked to actions, 
We use the human action datasets Kinetics-400~\citep{kay2017kinetics}, UCF-101~\citep{soomro2012ucf101}, and HMDB-51~\citep{kuehne2011hmdb} to circumvent the lack of quality annotations of motion descriptions. The advantage of action datasets is that for every action label, we can obtain the corresponding characteristic motion of the action by using large language models like  GPT-3~\cite{brown2020language} and GPT-4~\citep{openai2023gpt4}, which, to a large degree, produce accurate descriptions of the all the actions in the datasets. Figure \ref{fig:GPT4} shows some examples of the descriptions generated by GPT-4 and corresponding videos and original captions.

We evaluate several video-text models on the motion description retrieval task using the HMDB-51 and the UCF-101 datasets.  We compare against human performance and show that all models fall far behind, raising questions about the design of video-text models and the role of quality captions in training better models to capture human motion. To address this question, we propose a method described in section~\ref{method} to investigate if providing better motion description captions helps video-text models understand fine-grained motion descriptions. To validate this fairly, we compare our method with video-text models that similarly initialize their video encoder and text encoder with pre-trained CLIP~\citep{radford2021learning} weights so that undue performance gain is not obtained by pre-training on video-text data.  Our results show that our proposed pipeline is very effective in learning fine-grained motion descriptions on both the UCF-101 dataset and the HMDB-51 dataset.
In summary, our contributions are:
\begin{enumerate}
  \item Creating a dataset of human motion descriptions for three action datasets. 
  \item Evaluating current video-text models representing motion description in videos on the UCF-101 and HMDB-51 datasets against human expert evaluation.
  \item Introducing a method to validate the need for better captioning in video-text models to understand motion descriptions and demonstrate the method's effectiveness in capturing fine-grained motion descriptions.
\end{enumerate}

\section{ Background and related work}
\textbf{Multimodal datasets and video understanding tasks:}
Howto100M and Spoken moments in time are popular video caption datasets used in pre-training video-text models. ActivityNet, MSR-VTT, DiDeMo \citep{hendricks18emnlp}, VaTex \citep{wang2019vatex} are representative datasets used for video-language alignment tasks like video-to-text retrieval or text-to-video retrieval. To our knowledge, we are the first to introduce fine-grained motion descriptions in video datasets, which is not the focus of existing datasets. We introduce motion descriptions on Kinetics-400, HMDB-51, and UCF-101 datasets. 

\textbf{Video-text models}:
Various efforts have been made to build video-text models \citep{luo2022clip4clip,fang2021clip2video} mainly for video-to-text retrieval tasks. These models have developed mechanisms based on CLIP and extended them to video frames. Video-text models \citep{wang2021actionclip, ni2022expanding,wu2023revisiting, rasheed2023fine, Momeni_2023_ICCV}  have also been used for general video recognition both in the supervised and the zero-shot action recognition setting. \cite{Momeni_2023_ICCV} introduced verb-focused contrastive training to improve better verb reasoning by learning with hard negative verb examples. \cite{park2022exposing} introduced contrast sets to identify pitfalls in video-text models and recommended the need for fine-grained action understanding to tackle hard negatives in contrast sets. We differ from them as we utilize motion descriptions, which has its unique challenge. We compare our video-text model with Vanilla CLIP \citep{radford2021learning}, XCLIP \citep{ni2022expanding}, Text4Vis \citep{wu2023revisiting} and VifiCLIP \citep{rasheed2023fine}, primarily because they utilize the pre-trained CLIP for fair evaluation purposes.

\section{Benchmark} \label{bencmark}

\begin{figure*}[t!]

   \includegraphics[width=0.9\linewidth]{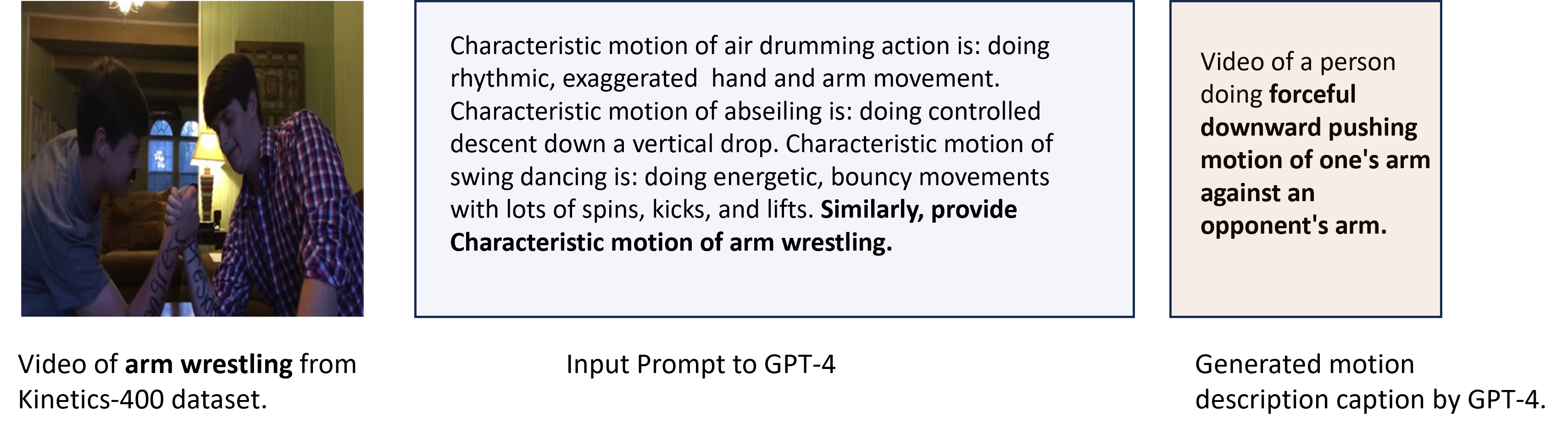}
  \centering
\caption{\textbf{ Schematic representation of the generation of motion descriptions in existing action datasets. }}
\label{fig:GPT4}
\end{figure*}

\textbf{Designing a motion description benchmark:} \label{motiond}
We perform experiments on Kinetics-400, UCF-101, and HMDB-51 datasets. For each class in these action datasets, we prompt GPT-4 to produce the characteristic motions of the action (given by the caption annotation). Figure \ref{fig:GPT4} shows the overall process of obtaining the motion descriptions. More details about the dataset statistics and generation are described in Appendix~\ref{secappendixD}.  The dataset can be accessed at \url{https://github.com/chinmayad/motiondescriptions.git}

We conducted a user study to evaluate the quality of generated motion descriptions. Following \citep{karpinska2021perils}, which questions the use of Amazon Mechanical Turk for such studies, we conducted this study with expert graduate student volunteers who have taken courses in computer vision and natural language processing. The evaluators were shown different questions in a training session. All the evaluators were trained with different questions and explained the project's overall goal and how they contributed to it. We asked two volunteer graduate students of different ages and ethnicities to participate in this study.

The following definitions were given.
\begin{enumerate}
\item Conciseness: Conciseness is related to the length and non-redundancy of the generated text. 
\item Hallucinations: Hallucinations are related to generating physically non-plausible motion descriptions.
\item Relevance: Relevance refers to how much correspondence there is between the objects, action, and motion description. 
\item Correctness: Correctness refers to how accurate the motion description is.
\item Harmfulness:  Is there any objectionable or harmful content in the generated motion description? 
\end{enumerate}
Each of the above attributes is evaluated on a 5-point Likert scale. We report the mean 5-point Likert score and IAA\%, the inter-annotator agreement that measures the percentage of descriptions where annotators gave the same rating. We asked the volunteers to rate the generated motion description with the above attributes for each motion description in each dataset. The study results are given in table\ref{table:quality}.

We noticed that many of the actions in the UCF-101 and HMDB-51 datasets involve objects, and the retrieval task is easier when an object is present in the generated motion description. For this reason, we also created another set of motion descriptions in which we replaced the names of objects with the generic word “object,” which made the task slightly more challenging.

\begin{table}[t]
\centering
\begin{tabular}{|c|c|c|c|}

\hline
Method & Mean & IAA\%  \\ 
\hline
Conciseness    & 3.86  & 47.5  \\
Hallucinations  & 1.12 &  19.35 \\
Relevance  & 3.4 &  87  \\
Correctness  & 3.92  & 72   \\
Harmfulness  & 1 & 100 \\
\hline
\end{tabular}
\caption{Evaluation of the quality of generated motion descriptions}
        \label{table:quality}

   %  \label{tab:temps}
\end{table}

\section{Proposed method}
\label{method}
\begin{figure*}[t!]

   \includegraphics[width=0.9\linewidth,height=0.43\textwidth]{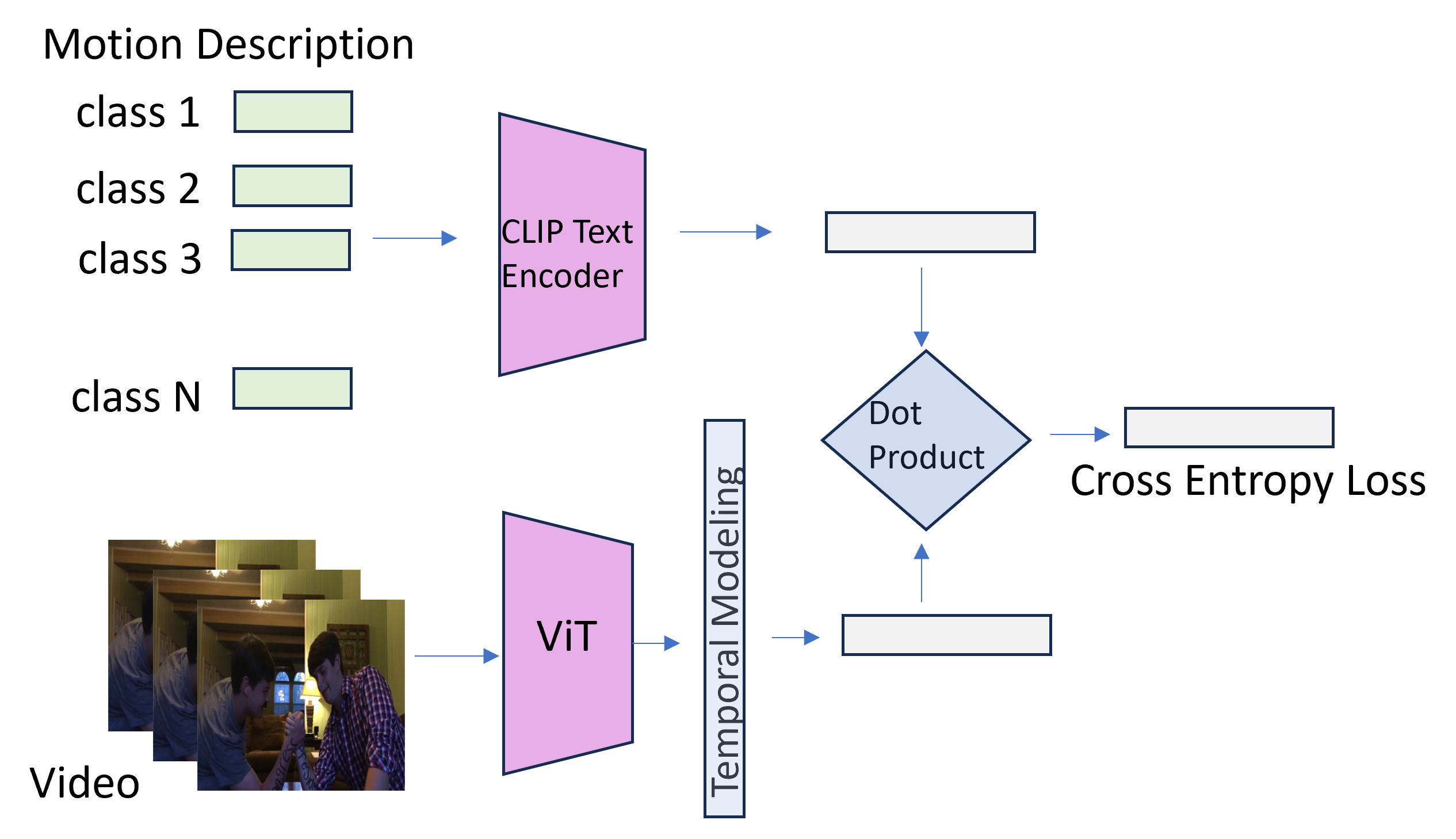}
  \centering
\caption{\textbf{ Schematic representation of our approach encoding motion description in video-text model pipeline:}  We integrate motion information as classifier weight in a supervised training paradigm. We finetune the image encoder to integrate the motion information while classifying videos in the kinetics400 dataset. }
\label{fig:overview}
\end{figure*}
This section describes our proposed video-text model that incorporates the textual motion description. A typical vision-language model like CLIP is usually trained using contrastive learning loss or its variations \citep{miech2020end,Momeni_2023_ICCV}. % cite Action verbs other contrastive learning methods. 
The quality of representations learned from the video-text model \citep{momeni2023verbs} will depend on the captions used during pre-training. Since no large video-caption dataset containing motion descriptions exists, we can't directly train a video-text model using contrastive learning. 
We, therefore, propose a method that utilizes the rich linguistical understanding of motion in actions from GPT-4 to give us the captions required for training.  Our approach has two parts: Generating the required motion descriptions of actions using GPT-4 described in section \ref{bencmark} and training video-text models to utilize these motion descriptions described in section~\ref{Model}.

\subsection{Problem setting}

The model is trained on kinetics-400 as source dataset $D_s$ and tested on target datasets $D_t$:  UCF101 and HMDB51.  
The source dataset $D_s$ consists of videos $x_s$ and labels $L_s$ belonging to classes $C_s$. The target dataset $D_t$ consists of videos $x_t$ and labels $L_t$ belonging to classes $C_t$. We use a zero-shot setting such that $C_s \cap C_t = \emptyset $. Let the generated motion descriptions from GPT-4 for $L_s$ and $L_t$ be $M_s$ and $M_t$ respectively.

\subsection{Architecture setting} \label{Model}
Figure \ref{fig:overview} gives an overview of our approach. While training, we freeze the text encoder from CLIP and fine-tune the visual encoder to learn the motion representation. While testing, motion descriptions are passed through the network to obtain classifier weights. Given a video from the target dataset, we obtain the logits indicating the probability that a video matches the corresponding motion description.  A detailed discussion on training, testing, and implementation details is given in Appendix~\ref{sec:appendixA}.   

\subsection{Theoretical justification}
We provide a theoretical justification for our proposed approach. Consider a large-scale dataset $D$ containing visual samples with ground-truth labels. Denoting our labeled dataset $D = {(x1, y1),(x2, y2), . . .},$ let $X$ represent input data ${x1,x2,...}$ and $Y$ represent the labels ${y1,y2,..}.$

A typical supervised learning framework with a linear predictor involves minimizing $L(X^TW, Y) + Sigma(W)$ where $W$ contains the parameters to be learned, $Sigma$ is the regularizing function, and $L$ is the loss function. Here, $W$ is learned independently on $D$ and will not be helpful for new classes or other downstream datasets. As proposed in \citep{romera2015embarrassingly}, to make the approach tractable for zero-shot learning, we need to make $W$ so it carries valuable information for new classes. The authors of \citep{romera2015embarrassingly} introduce 
\begin{equation} \label{eq:1}
W= V S^T,
\end{equation}
which we refer to as equation \ref{eq:1}, where $S$ is the signature of classes obtained from attributes of classes in $D$, and $V$ is a new set of parameters to be learned. 

For our scenario, we want to fine-tune the video-text model on the Kinetics dataset so that it can learn the motion description.

Let us denote $V_E$ as the visual encoder from CLIP or any pre-trained video-text model. The supervised learning formulation to learn $V_{E}^*$  and $W_{proj}^*$ as in \citep{wu2023revisiting} 
 can now be represented in minimizing cross-entropy $ H(y|\sigma(W_{proj} . V_E (x))) $    referred to as equation 2, where $H(p*|p)$ stands for the Cross Entropy between the predicted distribution $p$ and the ground-truth distribution $p*$. $\sigma$ denotes the softmax operation, $W_{proj} \in R^{c \times d}$ denotes the linear projection matrix for classification where $c$ is number of classes and $d$ is the dimension of embedding from $V_E$. The above formulation in equation 2 is a standard visual feature transferring paradigm, where the visual encoder $V_E$. and the projection matrix (classifier) $W_{proj}$ are learned simultaneously.  We need to introduce a motion description to make the formulation in equation 2 learn the motion description and be useful for recognizing new motion descriptions for zero-shot settings.

Inspired by equation 1 where $W=VS^T$,    we introduce motion description by making $W_{proj}$ the signature of classes $S$, and $V_E$ the new set of parameters of the visual encoder to be learned. In \citep{romera2015embarrassingly}  $S$ was obtained from an attribute matrix, and in our work, we obtain $W_{proj}$ as embeddings of a motion descriptor obtained from a CLIP text encoder.

\begin{table}[h]

\begin{tabular}{|c|c|c|c|}

\hline
Method & Object & Masked Object  \\ 
\hline
 Vanilla CLIP   &   25.92  & 23.33 \\
%XCLIP  &  &  \\
Text4Vis  & 51.23  & 33.80 \\

XCLIP  &  52.37 & 32.01 \\
ViFiCLIP  & 52.70 & 34.76 \\

%Vifi   &   &    \\ 34.76
Our Method  & 58.46 & 47.80 \\
\hline
Human estimate & 98 &98 \\
\hline
\end{tabular}
\caption{Evaluation of percentage accuracy in motion description retrieval task on UCF-101 dataset. }
\label{table:UCF101}

   %  \label{tab:temps}
\end{table}

\section{Results}

Task: For the target datasets UCF-101 and HMDB-51, the input is a video and the list of generated motion descriptions for all the classes in the dataset. The video-text model predicts the closest motion description that describes the video.  The metric used is the percentage accuracy of correctly predicted motion descriptions. 

The models we evaluate are Vanilla CLIP \citep{radford2021learning}, XCLIP \citep{ni2022expanding},
Text4Vis \citep{wu2023revisiting} and VifiCLIP \citep{rasheed2023fine}. %Vanilla CLIP consists of the pre-trained CLIP model with temporal averaging of frames. %XCLIP and 
%Text4Vis is trained on the Kinetics-400 dataset. 
Details about the models are given in Appendix~\ref{sec:appendixB}. 

Human estimated performance: We sampled five videos randomly from each class in UCF-101 and HMDB-51. A human expert was asked to select the motion description that correctly describes the video from the list of descriptions generated by GPT4.  Human experts are graduate students who have taken computer vision and NLP graduate courses and volunteered for this study. 

Table \ref{table:UCF101} reports the performance of various approaches on the UCF101 dataset.  Our proposed method beats previous methods by over 5\%  for motion descriptions containing the names of objects involved and by over 10\% for motion descriptions where the word ``object'' replaces the object's name.  We noticed that all the video-text models perform very poorly compared to human-estimated performance. We also see that video-text models have a strong bias toward nouns. When the specific name of the object involved is not used, there is an average  10\%  drop in performance for all methods, indicating the strong bias video-text models have for objects. Table \ref{table:HMDB} reports the performance of various approaches on the HMDB-51 dataset. Similar trends are found in the HMDB-51 dataset.

\begin{table}[t]

\begin{tabular}{|c|c|c|c|}

\hline
Method & Object & Masked Object  \\ 
\hline
 Vanilla CLIP   & 25.26  & 16.67  \\
XCLIP  & 29.35 &  19.35 \\
Text4Vis  & 34.12 &  24.93  \\
VifiCLIP  & 36.20  & 28.9   \\
Our Method  & 39.24 & 28.41 \\
\hline
Human estimate & 97.5 &96 \\
\hline
\end{tabular}
\caption{Evaluation of percentage accuracy in motion description retrieval task on HMDB-51 dataset.}
        \label{table:HMDB}

   %  \label{tab:temps}
\end{table}

\section{Conclusion}
We introduced a benchmark to understand how motion is understood in video-text models. We highlighted the limitations in obtaining quality annotations describing motion in video. We also showed that the performance of video-text models for retrieving motion descriptions is poor compared to human expert performance.  Our proposed method circumvents some of these issues and improves over other video-text models. While designing video-text models, we leave it to future work to build better models to capture motion and ignore the biases due to the object or the scene.

\section{Limitations}
We use a CLIP text encoder trained on image-text data to represent motion descriptions. This is not the best thing to do as, in practice, the CLIP text encoder would never have encountered the dynamics of videos while training. However, we hope the method works if we replace this CLIP text encoder with any other video-text model text encoder. Another limitation is that we trained on the Kinetics-400 dataset and tested on the UCF-101 and HMDB-51 datasets. As shown in the results section, the presence of an object or scene can often impact the performance during the retrieval task on these datasets. Furthermore, since we fine-tune image encoders on the source dataset, the possibility of overfitting the source dataset exists, leading to poor transferability on another target dataset. There are also potential biases in generated descriptions by GPT-4, and human quality estimation is expensive. For our experiments, volunteers spent a total of 16 hours.

\section{Ethical Considerations}
We aim to highlight the neglected aspect of modeling motion in video-text models. We think incorporating motion descriptions and reducing the biases of video-text models to objects and scenes positively impacts the design of video-text models. However, there could be a potential risk introduced in our method, as we rely on GPT-4 to provide us with motion descriptions of actions. 
\section{Acknowledgements}
The support of NSF under awards OISE 2020624, BCS 2318255, and ARL under the Army Cooperative Agreement W911NF2120076 is greatly acknowledged.
\bibliography{custom}
\newpage
\appendix

\section{Implementation details of our method}
\label{sec:appendixA}
\subsection{Problem Setting:}

The model is trained on kinetics-400 as source dataset $D_s$ and tested on target datasets $D_t$:  UCF-101 and HMDB-51.  
The source dataset $D_s$ consists of videos $x_s$ and labels $L_s$ belonging to classes $C_s$. The target dataset $D_t$ consists of videos $x_t$ and labels $L_t$ belonging to classes $C_t$. We use a zero-shot setting such that $C_s \cap C_t = \emptyset $. Let the generated motion descriptions from GPT-4 for $L_s$ and $L_t$ be $M_s$ and $M_t$ respectively.

\subsection{Training}
Motion descriptions $M_s$ are passed through a frozen CLIP text encoder to obtain the class prototypes of $L_s$. Our intuition is that these class prototypes can be approximated as classifier weights of the supervised video classifier. The concept takes its motivation from the work of \cite{nukrai2022text,liang2022mind}, which shows that text embeddings from CLIP and vision embeddings from CLIP are very similar and fall within a ball of small radius. The obtained CLIP embeddings of motion descriptions from the frozen text encoder would approximately translate to visual class prototypes if obtained visually.  With that intuition, we use the class prototypes from the frozen CLIP text encoder as the classifier weights of a visual classifier. 

Given a video $v_s$ from the source dataset $D_s$, T frames are sampled uniformly. The sampled frames are passed through a CLIP pre-trained Image encoder and temporally pooled to obtain a visual feature of the video. Then, the logits are obtained by computing the dot product of this video feature with the transpose of the classifier weights $W_s$. The model is trained using cross-entropy loss over logits and labels $L_s$, and the parameters of the CLIP image encoder are updated.

\subsection{Testing}
The motion descriptions $M_t$ are passed through the network to obtain classifier weights  $W_t$. Given a video $v_t$ from the target dataset $D_t$, we obtain the logits indicating the probability that a video matches the corresponding motion description $M_t$. 

\subsection{Experimental details}

Our model uses a VIT-B/16 pre-trained CLIP text and image encoder. We use eight frame samples per video. The CLIP text encoder was kept frozen during the training, and the CLIP image encoder was fine-tuned. 
We train the model for $10$ epochs on the Kinetics-400 dataset with a learning rate of $0.00005$ with a batch of the size of $20$ on $4$ NVIDIA RTX A5000 for 40 GPU hours.  We use a learning warm step of $5$ and a weight decay of 0.2. We use the Adam optimizer with the cross-entropy loss for training on the Kinetics dataset with a clip ratio of $0.1$. Here, we describe more details about our baselines. We report the best results after running experiments on $5$ runs. 

\subsection{Temporal modeling}
We experimented with adding a 6-layer temporal transformer on the video head of the VIT-B/16 transformer. The results are shown below in table \ref{table:temporal} and \ref{table:temporal2}.  Contrary to our initial hypothesis, having a temporal transformer didn’t improve the performance over mean average pooling.

\begin{table}[h]

\begin{tabular}{|c|c|c|}

\hline
\makecell{Method} & \makecell{Object} & \makecell{Masked Object}  \\ 
\hline
\makecell{Temporal \\ transformer}   &  \makecell{57.02} & \makecell{44.4} \\
\hline
\makecell{Mean Average \\ Pooling}  & \makecell{58.46} & \makecell{47.80} \\

\hline
\end{tabular}
\caption{Evaluation of percentage accuracy in motion description retrieval task on UCF-101 dataset. }
\label{table:temporal}

   %  \label{tab:temps}
\end{table}

\begin{table}[h]

\begin{tabular}{|c|c|c|}

\hline
Method & Object & Masked Object  \\ 
\hline
\makecell{Temporal \\ transformer} & \makecell{37.53}   & \makecell{26.84}  \\
\hline
\makecell{Mean Average \\ Pooling}  &  \makecell{39.24} & \makecell{28.41} \\

\hline
\end{tabular}
\caption{Evaluation of percentage accuracy in motion description retrieval task on HMDB-51 dataset. }
\label{table:temporal2}

   %  \label{tab:temps}
\end{table}

\subsection{Does fine-tuning cause overfitting?}
As in any fine-tuning method, there is a risk of overfitting the source dataset. We performed experiments to see if overfitting is an issue. Based on our experiments, the degree to which the model overfits is negligible compared to the method's overall improvement. Table \ref{table:ft1} and Table \ref{table:ft2} below show the accuracies at different epochs of fine-tuning the vision encoder. 

\begin{table}[h]

\begin{tabular}{|c|c|c|}

\hline
\makecell{Number of Epochs} & \makecell{Object} & \makecell{Masked Object}  \\ 
\hline
\makecell{Epoch 5}   &  \makecell{57.50} & \makecell{46.59} \\
\hline
\makecell{Epoch 10}  & \makecell{58.46} & \makecell{47.80} \\

\hline
\makecell{Epoch 20}  & \makecell{58.2} & \makecell{47.02} \\

\hline

\end{tabular}
\caption{Evaluation of percentage accuracy in motion description retrieval task on UCF-101 dataset. }
\label{table:ft1}

   %  \label{tab:temps}
\end{table}

\begin{table}[h]

\begin{tabular}{|c|c|c|}

\hline
Number of Epochs & Object & Masked Object  \\ 
\hline
\makecell{Epoch 5} & \makecell{38.25}   & \makecell{27.82}  \\
\hline
\makecell{Epoch 10}  &  \makecell{39.239} & \makecell{28.41} \\

\hline
\end{tabular}
\caption{Evaluation of percentage accuracy in motion description retrieval task on HMDB-51 dataset. }
\label{table:ft2}

   %  \label{tab:temps}
\end{table}

\section{ Baselines}
\label{sec:appendixB}

\subsection{Vanilla CLIP:}

We use VIT-B/16 pre-trained CLIP text and image encoder obtained from \citep{radford2021learning}  and temporally average the frame outputs while evaluating HMDB-51 and UCF-101 datasets. 

\subsection{XCLIP}
We use  the XCLIP \citep{ni2022expanding} model and code available from  \url{https://huggingface.co/docs/transformers/model_doc/xclip}. We use "microsoft/xclip-base-patch16-zero-shot" model from huggingface.co while evaluating the HMDB-51 and UCF-101 datasets. 
%\subsection{ XCLIP:}
%We use hugging face model of XCLIP zero-shot. 

\subsection{Text4Vision}
We use the VIT-B/16 base architecture with $8$ frames per video. We use pre-trained weights from \url{https://github.com/whwu95/Text4Vis/tree/main}

\subsection{VifiCLIP}
We use the VIT-B/16 architecture and obtain the model and code from the official implementation of VifiCLIP \citep{rasheed2023fine} from \url{https://github.com/muzairkhattak/ViFi-CLIP}.

\section{Dataset}
\subsection{Kinetics-400}
Kinetics-400 is a large-scale dataset containing 400 classes downloaded from YouTube. It has 240K training videos and 20K validation videos. Some videos are missing if the YouTube user has removed them. 

\subsection{UCF-101}
 The UCF-101 human action dataset consists of 13 K YouTube videos belonging to 101 classes. We report results on full classes on one split provided by the authors. 

\subsection{HMDB-51 }
 It contains approximately 7K videos belonging to 51 classes. We report results on the split provided by the authors.

\section{Motion Description Generation}
\label{secappendixD}
We use the GPT -4 API to obtain motion descriptions. The generated motion descriptions for the three datasets are provided as supplementary data along with this submission.

We noticed from experiments that we needed to provide some example motion descriptions in the prompt to obtain motion descriptions in the format we are interested in. The prompt we used is \enquote{\textit{Characteristic motion of air drumming action is: doing rhythmic, exaggerated hand and arm movement.  The characteristic motion of abseiling is:  doing a controlled descent down a vertical drop. The characteristic motion of swing dancing is: doing energetic, bouncy movements with lots of spins, kicks, and lifts. Similarly, provide the characteristic motions of {action X}.}}. %Example motion descriptions generated are shown in the Appendix.

\subsection{GPT-4 generated motion descriptions quality control}  5 volunteers evaluated the generated motion descriptions for pairwise comparison. The pairwise comparison included selecting the best one among two generated motion descriptions. A vote of majority was used to select the final motion description. Volunteers with diverse experience and age groups were selected to reduce bias. 

\subsection{Dataset statistics}
The kinetics-400 dataset consists of 400 classes, the UCF-101 human action dataset consists of 101 classes and HMDB-51 consists of 51 classes. We generate a characteristic motion description for each class in all three datasets. For UCF101 and HMDB-51, we mask objects manually after obtaining the motion description. Table \ref{table:st} provides the motion description dataset statistics.

\begin{table*}[t!]
    \centering
    \begin{tabular}{|c|c|c|c|c|c|}\hline

        \hline
        \makecell{\textbf{Dataset}} & \makecell{Number of \\ videos}& \makecell{Number of \\ unique motion \\descriptions} & \makecell{Average number \\ of verbs per \\ motion description } & \makecell{Average number \\ of words per \\ motion description}& \makecell{ Number of \\ verbs in the \\ motion descriptions}\\
        \hline
        \makecell{Kinetics 400} & \makecell{246000}& \makecell{400}&\makecell{3.4}&\makecell{19} &\makecell{1371}\\
        \hline
        \makecell{UCF 101} & \makecell{13320}& \makecell{ 101}&\makecell{3.2 }&\makecell{19}&\makecell{325}\\
                \hline
        \makecell{HMDB 51} & \makecell{6849}& \makecell{51}&\makecell{3.2 }&\makecell{17}&\makecell{164}\\
        \hline
    \end{tabular}
    \caption{ Statistics of motion description dataset}
\end{table*}
\label{table:st}
\end{document}